%% file: acl2023.tex
\pdfoutput=1

\documentclass[11pt]{article}

\usepackage{ACL2023}

\usepackage{times}
\usepackage{latexsym}
\usepackage[normalem]{ulem}
\usepackage{enumitem}

\usepackage{rycolab}
\usepackage{hexatagger}

\usepackage[T5,T1]{fontenc}

\usepackage[utf8]{inputenc}

\usepackage{microtype}

\usepackage{inconsolata}

\title{Hexatagging: Projective Dependency Parsing as Tagging}

\author{Afra Amini\thanks{\hspace{0.5em}Equal contribution.} \qquad Tianyu Liu\footnote[1]{} \qquad Ryan Cotterell \\
\setlength{\fboxsep}{2.5pt}%
\setlength{\fboxrule}{2.5pt}%
\fcolorbox{white}{white}{
    $\{$\texttt{\href{mailto:afra.amini@inf.ethz.ch}{afra.amini}, }\texttt{\href{mailto:tianyu.liu@inf.ethz.ch}{tianyu.liu}, }\texttt{\href{mailto:ryan.cotterell@inf.ethz.ch}{ryan.cotterell}}$\}$\texttt{@inf.ethz.ch}
} \\
    {%
\setlength{\fboxsep}{2.5pt}%
\setlength{\fboxrule}{2.5pt}%
\fcolorbox{white}{white}{
    \includegraphics[width=.15\linewidth]{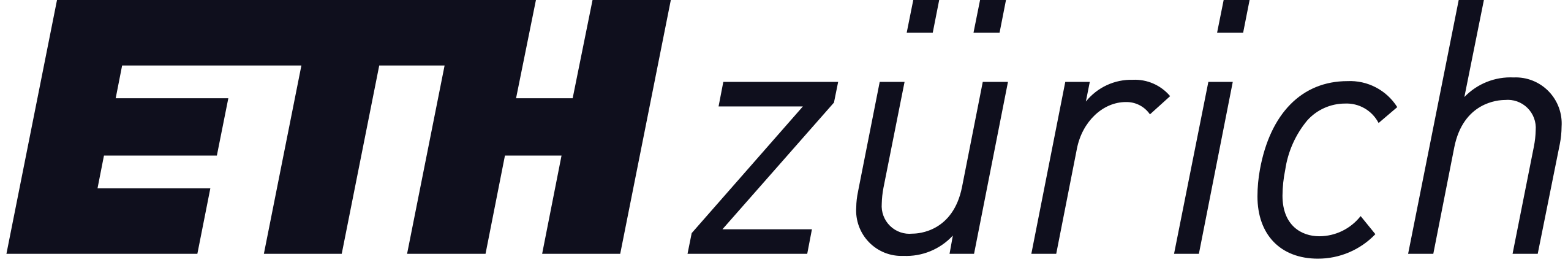}
}
}}

\begin{document}
\maketitle

\begin{abstract}
We introduce a novel dependency parser, the hexatagger, that constructs dependency trees by tagging the words in a sentence with elements from a \emph{finite} set of possible tags. In contrast to many approaches to dependency parsing, our approach is fully parallelizable at training time, i.e., the structure-building actions needed to build a dependency parse can be predicted in parallel to each other. Additionally, exact decoding is linear in time and space complexity. Furthermore, we derive a probabilistic dependency parser that predicts hexatags using no more than a linear model with features from a pretrained language model, i.e., we forsake a bespoke architecture explicitly designed for the task. Despite the generality and simplicity of our approach, we achieve state-of-the-art performance of 96.4 LAS and 97.4 UAS on the Penn Treebank test set. Additionally, our parser's linear time complexity and parallelism significantly improve computational efficiency, with a roughly 10-times speed-up over previous state-of-the-art models during decoding.
\newline
 \newline
 \vspace{0.2em}
  \hspace{.5em}\includegraphics[width=1.25em,height=1.25em]{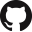}\hspace{.75em}\parbox{\dimexpr\linewidth-2\fboxsep-2\fboxrule}
  {\url{https://github.com/rycolab/parsing-as-tagging}}
\end{abstract}

\section{Introduction} 
The combination of parallel computing hardware and highly parallelizable neural network architectures \cite{attention} has enabled the pretraining of language models on increasingly large amounts of data.
In order to apply pretrained language models to downstream NLP tasks, many practitioners fine-tune the pretrained model while the task-specific architecture is jointly trained from scratch.
Typically, the task-specific architecture is built upon the hidden representations generated by the final layer of a pretrained model. 
Exploiting pretrained language models in this manner has boosted the performance considerably on many NLP tasks \cite{bert, electra, aghajanyan-etal-2021-muppet}.
However, for the end-to-end fine-tuning process to be fully parallelizable, it is also necessary to parallelize the training of the task-specific architecture.
Unfortunately, due to the complexity of the output in many structured prediction tasks in natural language, e.g., in dependency parsing, state-of-the-art models still use architectures with limited parallelization during training \cite{mrini-etal-2020-rethinking,yang-tu-2022-headed}.\looseness=-1 

\begin{figure}[t]
    \centering
     \setlength{\belowcaptionskip}{-10pt}
\includegraphics[width=\columnwidth]{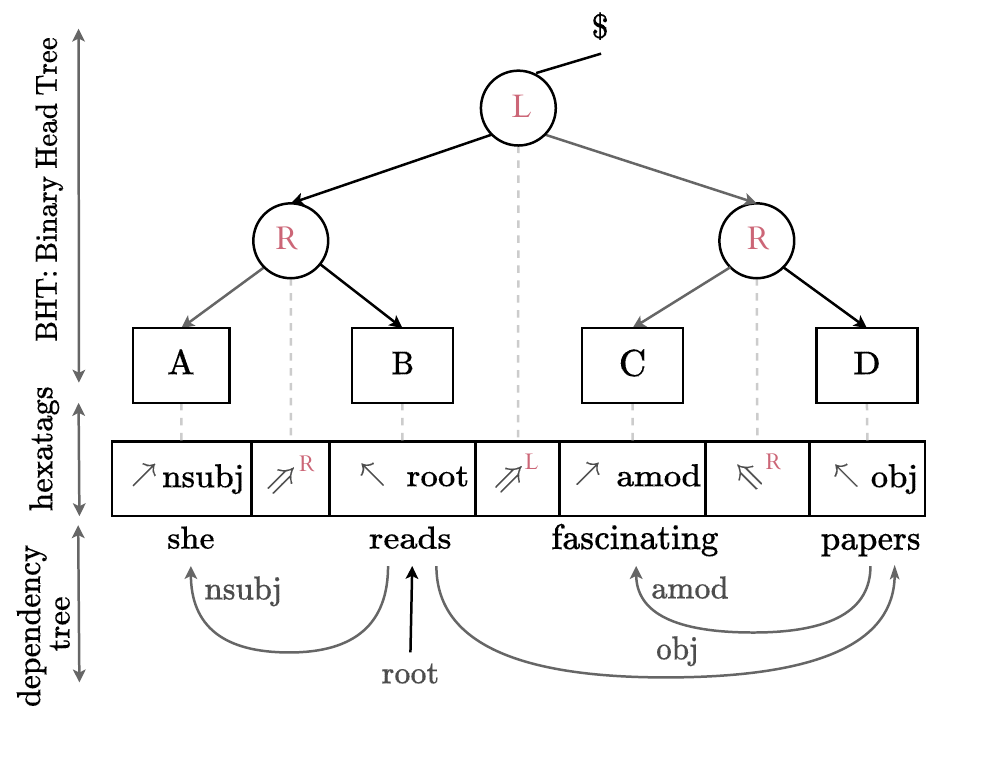}
    \caption{From bottom to top, the figure shows the dependency tree, the hexatags, and the binary head tree for the sentence ``She reads fascinating papers.''}
    \label{fig:example}
\end{figure}
In an attempt to develop parsers parallelizable during training, a recent line of work recasts parsing as tagging \citep{li-etal-2018-seq2seq, strzyz-etal-2019-viable, kitaev-klein-2020-tetra, afra-parsing}. 
Under this approach, a parse tree is linearized into a sequence of tags.\footnote{In some tagging-based dependency parsers, the cardinality of the set of tags even grows as a function of the length of the input sequence and, thus, is unbounded.}
The benefit of such a paradigm is that tagging can be done by only adding a linear classifier on top of a pretrained language model and the tags can, thus, be predicted independently.
This leads to a parser that is highly parallelizable and whose training can be easily harmonized with the (parallelizable) fine-tuning of pretrained language models.
During decoding, an exact algorithm is used to recover a valid sequence of tags which is then converted back to a parse tree.\looseness=-1

\citet{kitaev-klein-2020-tetra} were the first to propose a parsing-as-tagging scheme with a constant tag space for \emph{constituency parsing} and, additionally, the first to achieve results competitive with the state-of-the-art non-parallelizable constituency parsers using such a tagger.
However, for dependency parsing, all dependency parsing-as-tagging schemes in the literature \cite{li-etal-2018-seq2seq,strzyz-etal-2019-viable,vacareanu-etal-2020-parsing} have infinite tag sets whose cardinality grows with the length of the input sequence, which limits such parsers' efficiency and generality \cite{strzyz-etal-2019-viable}.
Moreover, in some cases, this growth hinders generalization to sentences longer than the longest training sentence.
Furthermore, tagging-based dependency parsers still do not perform competitively with the best-performing parsers in the literature \cite{li-etal-2018-seq2seq}.\looseness=-1

In this paper, we propose a novel way of framing projective dependency parsing as a tagging task.
Our approach makes use of 6 distinct tags, motivating us to naming the scheme \defn{hexatagger}.
In our experiments, hexatagger achieves state-of-the-art performance on the English Penn Treebank \cite[PTB;][]{marcus-etal-1993-building} test set.
Notably, it outperforms parsers with more computationally expensive training procedures and extra constituency annotations, e.g., the parser developed by \citet{mrini-etal-2020-rethinking}.
Furthermore, hexatagger achieves results competitive to \citeposs{yang-tu-2022-headed} parser on the Chinese Penn Treebank \cite[CTB;][]{ctb} test set and 12 languages on the pseudo-projectivized data from the Universal Dependencies \cite[UD2.2;][]{ud2.2} benchmark.
In terms of efficiency, our experiments suggest that hexatagger is 10 times faster than previous top-performing parsers, and consumes significantly less memory, despite using an exact dynamic program for decoding.\looseness=-1

\section{Hexatagging} 

In this section, we introduce hexatagging, a tagging scheme that consists of 6 unique tag types. We further prove by construction that there exists an injective mapping between valid sequences of hexatags and dependency trees.

\subsection{Binary Head Trees}\label{sec:tree2tag}
Before going into the details on how to represent dependency trees with a sequence of tags, we introduce \defn{binary head trees} (BHTs), a simple formalism that serves as a useful intermediary between dependency trees and sequence of hexatags.
Intuitively, a BHT is a special form of a constituency tree where each internal node is either labeled \circledl{} when the head of the derived constituent is in the left subtree 
or \circledr{} when the head is in the right subtree. 
See \Cref{fig:example} for a visual depiction of a BHT. In the next theorem, we formally state the relationship between the set of dependency trees and BHTs.\looseness=-1

\begin{theorem} \label{thm:existence-bht}
There exists a bijective\footnote{We remark that the bijectivitiy follows from a canonical ordering (left-to-right and inside-out) of a node's dependents.} function that maps every projective dependency tree to a BHT.
\end{theorem}
In the following two paragraphs, we sketch a construction that such a function exists, i.e., we describe how to map any dependency tree to a BHT and then how to map back any BHT to a dependency tree and back again.\looseness=-1

\paragraph{Projective Dependency Trees to BHTs.} To convert a dependency tree to a BHT, we start from the root and do a depth-first traversal of the dependency tree.
To avoid spurious ambiguity \citep{eisner-satta-1999-efficient}, we canonically order arcs of the tree by processing the arcs left to right and inside out.\footnote{One can also process the right arcs first. 
In our experiments, however, we observed no significant difference in the performance of the parser, see \Cref{app:left-right} for more analysis.}
Algorithmically, converting a dependency tree to a BHT proceeds as follows. When we first visit a word, we push it onto a stack and proceed with visiting its dependents. 
When there is no dependent word left to visit, we create a new node (\circledl{} or \circledr{}) and attach the top two elements in the stack as the left and right child of this node.
A step-by-step demonstration of this algorithm is shown in \Cref{tab:dep2tree} and pseudocode is provided in \Cref{alg:dep2bht}.

\paragraph{BHTs to Projective Dependency Trees.} To convert a BHT back to the dependency tree we follow \Cref{alg:bht2dep}. Algorithmically, we process BHT in a depth-first fashion. Upon visiting \circledr{} or \circledl{} nodes, we combine the top two elements in the stack by creating a dependency arc between them. The direction of the arc is determined by the label of the node (\circledr{} or \circledl{}). 
See \Cref{tab:tag2tree} for an example.
\input{figs/dep2tree.tex}
\newcommand{\pushstack}[1]{$\textsc{Push}\left(\text{#1}\right)$}
\newcommand{\makerightarc}[1]{$\textsc{MakeNode}\left(\text{#1}\right)$}
\newcommand{\makeleftarc}[1]{$\textsc{MakeNode}\left(\text{#1}\right)$}

\begin{figure}[!ht]
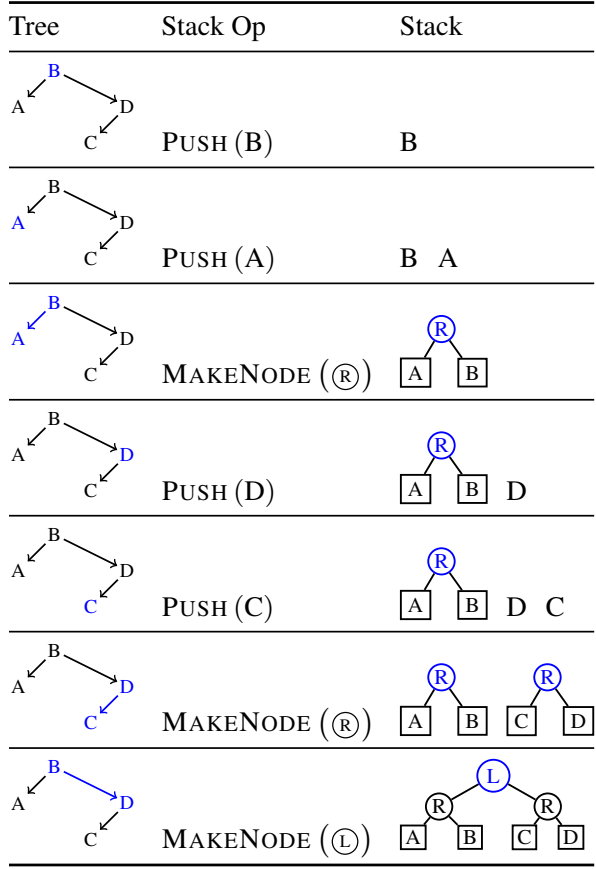

  \small
 \setlength{\belowcaptionskip}{-8pt}
  \begin{center}
  \resizebox{\columnwidth}{!}{\begin{tabular}{@{}l@{\hskip 8pt}l@{\hskip 8pt}l@{}}
    \toprule
    Tree & Stack Op & Stack \\
    \midrule
     \scalebox{0.7}{\depb} & \pushstack{B}     & B   \\ \midrule
     \scalebox{0.7}{\depa} & \pushstack{A}     & B \, A   \\ \midrule
     \scalebox{0.7}{\depab}& \makeleftarc{\circledr{}}    & \scalebox{0.75}{\rgilas{A}{B}}   \\ \midrule
     \scalebox{0.7}{\depd} & \pushstack{D}     & \scalebox{0.75}{\rgilas{A}{B}} \, D \\ \midrule
     \scalebox{0.7}{\depc} & \pushstack{C}     & \scalebox{0.75}{\rgilas{A}{B}} \, D \, C \\ \midrule
    \scalebox{0.7}{\depdc} & \makeleftarc{\circledr{}}    & \scalebox{0.75}{\rgilas{A}{B}} \, \scalebox{0.75}{\rgilas{C}{D}} \\ \midrule
    \scalebox{0.7}{\depbd} & \makerightarc{\circledl{}}   & \scalebox{0.75}{\dogilas} \\
    \bottomrule
  \end{tabular}}
  \end{center}
  \caption{\label{tab:dep2tree} \vspace{-1ex}The example shows how to derive the BHT for a dependency tree \scalebox{0.75}{\depfull}. The top of the stack is on the right.}
\end{figure}
\noindent Once the dependency tree is converted to a BHT, we can linearize it to a sequence of hexatags in a straightforward manner. 
\Cref{thm:bht} states the relationship between BHTs and hexatags formally.\looseness=-1
\begin{theorem} \label{thm:bht}
There exists a total and injective function that maps every BHT to a valid hexatag sequence, i.e., in other words, every BHT can be mapped to a \emph{unique} hexatag sequence.
However, some hexatag sequences do \emph{not} correspond to BHTs, i.e., the function is not surjective.\looseness=-1
\end{theorem}
In the following subsections, we prove by construction that such a function exists.
Throughout the rest of the paper, we refer to those haxatagging sequences that \emph{do} correspond to BHTs as \defn{valid}.
\subsection{From BHT to Hexatags}
To transform a given BHT to a sequence of hexatags, we enumerate the action sequence that a left-corner shift--reduce parser would take when parsing this BHT \citep{johnson-1998-finite-state}.
Left-corner parsers have actions that align more closely with the input sequence than top-down or bottom-up shift--reduce actions and, thus, offer a better linearization for tagging tasks \citep{afra-parsing}.
A simple explanation of this linearization process is given by \citet[\S 3.1]{kitaev-klein-2020-tetra}.  
Their algorithm involves an in-order traversal of the tree.
Upon visiting each node, we generate a tag that includes the direction of the arc that attaches the node to its parent, i.e., whether that node is a left or a right child of its parent, and the label of the node.
When traversing a BHT, this paradigm results in 6 distinct tag types:
\setlist{nolistsep}
\begin{itemize}[leftmargin=*]\setlength{\itemsep}{0pt} \setlength{\parskip}{0pt} \setlength{\parsep}{0pt}
    \item $\rightchild$: this terminal node is the right child of its parent;
    \item $\leftchild$: this terminal node is the left child of its parent;
    \item $\Rightchildr$ ($\Rightchildl$): this non-terminal node is the right child of its parent and the head of the corresponding constituent is on the right (respectively, left) subtree; 
    \item $\Leftchildr$ ($\Leftchildl$): this non-terminal node is the left child of its parent and the head of the corresponding constituent is on the right (respectively, left) subtree.
\end{itemize}
For an input sequence $\bw=w_1 \cdots w_N$, this process gives us a hexatag sequence of length $2N-1$. \Cref{fig:example} depicts tree-to-tags transformation through an example.

\paragraph{Labeled Dependency Trees.} 
When converting a \emph{labeled} dependency tree to a sequence of hexatags, the arc labels must be encoded in the tags. Therefore, while reading a terminal node, we concatenate the label of the arc that connects the node to its parent with the hexatag. 
In this case, the number of distinct tags would be $\bigo{|\calA|}$, where $|\calA|$ is the number of unique arc labels.
For example, in \Cref{fig:example} the hexatag generated while processing \textit{she} is: $\langle \leftchild, \texttt{nsubj}\rangle$.

\subsection{From Hexatags to Dependency Trees} \label{sec:tag2tree}
\begin{figure}[!ht]
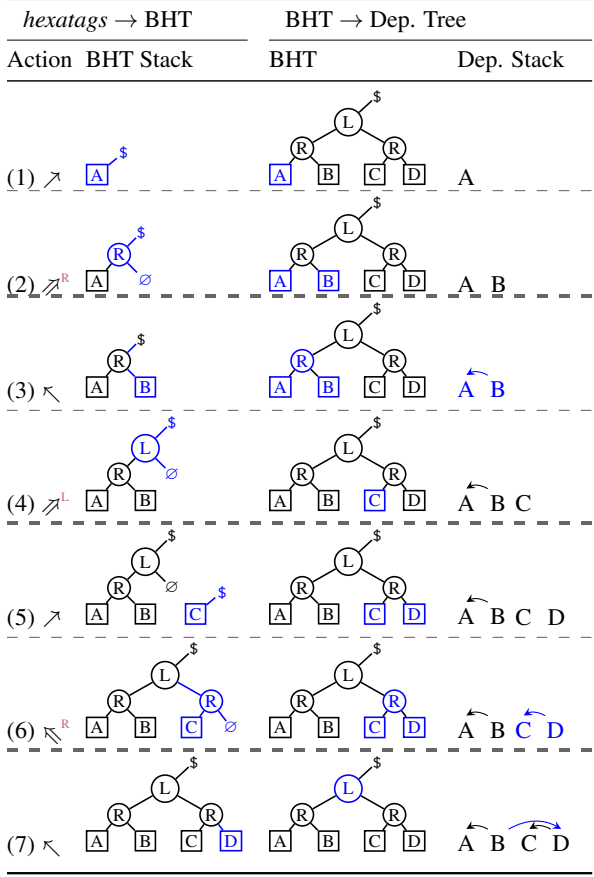

  \small
 \setlength{\belowcaptionskip}{-5pt}
  \begin{center}
  \resizebox{\columnwidth}{!}{%
  \begin{tabular}{@{}l@{\hskip 5pt}l@{\hskip 10pt}l@{\hskip 12pt}l@{}}
    \toprule
    \multicolumn{2}{l}{\emph{hexatags} $\rightarrow$ BHT} & \multicolumn{2}{l}{BHT $\rightarrow$ Dep. Tree} \\
    \cmidrule(l{0pt}r{8pt}){1-2}    \cmidrule(l{0pt}r{0pt}){3-4} 
    Action & BHT Stack & BHT & Dep. Stack \\
    \midrule
    (1) $\leftchild$  & \scalebox{0.75}{\newstacka{A}}  &   \scalebox{0.75}{\bhta}      & A  
    \mydashrule
    (2) $\Leftchildr$ & \scalebox{0.75}{\newstackaa}    &   \scalebox{0.75}{\bhtb}      & A \, B 
    \mythickdashrule
    (3) $\rightchild$  & \scalebox{0.75}{\newstackaab}  &   \scalebox{0.75}{\bhtR}      & \scalebox{1}{\depdobluel{A}{B}}
    \mydashrule
    (4) $\Leftchildl$ & \scalebox{0.75}{\newstackaabb}  &  \scalebox{0.75}{\bhtc}       & \depdol{A}{B} C 
    \mythickdashrule
    (5) $\leftchild$  & \scalebox{0.75}{\newstackaabbb} \scalebox{0.75}{\newstacka{C}}  &  \scalebox{0.75}{\bhtd} & \depdol{A}{B} C \, D \mydashrule
    (6) $\Rightchildr$  & \scalebox{0.75}{\newstackaabbc} &  \scalebox{0.75}{\bhtRR}    & \depdol{A}{B}  \depdobluel{C}{D}    \mythickdashrule
    (7) $\rightchild$  & \scalebox{0.75}{\newstackaabbcd} &  \scalebox{0.75}{\bhtL}     & \depfullblue   \\
    \bottomrule
  \end{tabular}}
  \end{center}
  \caption{\label{tab:tag2tree} The example shows how to derive BHT from hexatags and how to transform BHT to a dependency tree. The top of the stacks is on the right.}
  \vskip -.12in
\end{figure}

To transform a sequence of hexatags back to a dependency tree, we again go through a two-step process. 
First, we again interpret hexatags as actions in a left-corner shift--reduce transition system to construct a BHT. 
The actions in such a transition system are as follows: \looseness=-1
\setlist{nolistsep}
\begin{itemize}[leftmargin=*]\setlength{\itemsep}{0pt} \setlength{\parskip}{0pt} \setlength{\parsep}{0pt}
    \item $\leftchild$: shift the leaf node into the stack;
    \item $\Leftchildr$  ($\Leftchildl$): create a new node labeled $\circledr{}$ (respectively, $\circledl{}$), attach the top element in the stack as its left child, and attach a dummy node as its right child ($\varnothing$ in step 2 in \Cref{tab:tag2tree});
    \item $\rightchild$: pop the subtree on the top of the stack. Replace the dummy node in the subtree with the terminal node. Push the subtree back to the stack;
    \item $\Rightchildr$ ($\Rightchildl$): create a new node labeled $\circledr{}$ (respectively, $\circledl{}$). Pop the top element of the stack, attach it as the new node's left child, and set a dummy node as the node's right child. Pop another subtree of the stack, identify the dummy node in the subtree and replace it with the newly created subtree. Push the subtree back to the stack (step 6 in \Cref{tab:dep2tree});
\end{itemize} \vspace{-1ex}

\section{Probability Model} \label{sec:model}
In this section, we explain how to predict hexatags in parallel. Our tagging model predicts two hexatags for each word in the input sequence with the exception of that last word, for which we only predict one tag.
As discussed in \Cref{sec:tree2tag}, a hexatagger produces a sequence of $2N-1$ tags $\bt = [t_1, t_2, \dots, t_{2N-1}]$ for an input sequence of length $N$,  $\bw = w_1 w_2 \cdots w_N$. Therefore, an intuitive way to match the tag sequence with the input sequence is to assign two tags to each word.
We denote a training corpus $\corpus$ of $M$ tuples of input sequences and tag sequences $\{(\bw^m, \bt^m)\}_{m=1}^M$.\looseness=-1

To learn the scoring function over tags, we follow the same independence assumption as in \citep{kitaev-klein-2020-tetra}, i.e., the probability of predicting each tag is independent of other tags given the input sequence. 
This assumption barely harms model performance \citep[see][Table 3]{afra-parsing}, but significantly speeds up the training process by enabling each tag to be predicted in parallel, and complexity reduces by a factor of $\bigo{N}$.
The training objective is to minimize the negative log-likelihood of the gold-standard tag sequences, i.e.\looseness=-1
\begin{subequations}
    \begin{align}
        \loss(\param) &= - \sum_{(\bw, \bt) \in \corpus} \log\, p_{\param}(\bt \mid \bw) \\
        &= - \sum_{(\bw, \bt) \in \corpus} \log\, \prod_{n=1}^{2N-1} p_{\param}(t_n \mid \bw) \\
        &= - \sum_{(\bw, \bt) \in \corpus} \biggl( \sum_{n=1}^{N} \log\, p_{\param}(t_{2n-1} \mid \bw) \\
        & \qquad \qquad  +  \sum_{n=1}^{N-1} \log\, p_{\param}(t_{2n} \mid \bw) \biggr) \nonumber
    \end{align}
\end{subequations}
where $\param$ refers collectively to the parameters of the two linear projections and the parameters of the pretrained model.
\input{tabs/ud.tex}
\input{tabs/ptb.tex}
To obtain $p_{\param}(t_{2n} \mid \bw)$ and $p_{\param}(t_{2n+1} \mid \bw)$, we apply two independent linear projections on the contextualized representation of $w_n$\footnote{If a word consists of more than one subword, we apply the projection to the last subword.} given by a pretrained model and convert that to a probability distribution using softmax.

\section{Decoding}
Our goal in this section is to develop an efficient algorithm to find the highest-scoring hexatag sequence under the model developed in \Cref{sec:model}. 
As stated in \Cref{thm:bht}, the transformation function between BHTs and hexatag sequences is not surjective, i.e., not all the tag sequences can be transformed back into a BHT.
Therefore, we need to find a \emph{valid} hexatag sequence with the maximum probability under the model that \emph{can} be transformed back to a BHT. Once such hexatag sequence is found, we can follow the two-step algorithm described in \Cref{sec:tag2tree} to obtain the corresponding dependency tree.

To find the highest-scoring valid hexatag sequence, we follow the linear-time algorithm developed by \citet{kitaev-klein-2020-tetra}. 
For a hexatag sequence to be valid, we should be able to interpret it as actions in a left-corner shift--reduce transitions system, described in \Cref{sec:tag2tree}.
Concretely:
\setlist{nolistsep}
\begin{itemize}
    \item The first action can only be $\leftchild$ because other actions need at least one item in the stack;
    \item The actions $\Rightchildl, \Rightchildr$ can only be performed if there is at least two items in the stack;
    \item After performing all the actions, the stack should contain a single element. 
\end{itemize}
The above shows that the validity of a hexatag sequence only depends on the \emph{number} of elements in the stack at each point of the derivation.\footnote{
Specifically, The decoding algorithm can be thought of as constructing a lattice where each node corresponds to the number of elements in the stack for each transition step ($N \times d$ nodes for maximum stack size of $d$, $d \leq N$).
Each transition corresponds to performing a valid action. 
The score of the tag at step $n$ is set to the negative log probability $-\log\, p_{\param}(t_{n} \mid \bw)$ of the corresponding hexatag given by the model.
Finally, we remark that our decoding algorithm is essentially a shortest-path dynamic program that finds the highest-scoring valid hexatag sequence.
See \citet[\S 5.1]{afra-parsing} for a deeper discussion of this point.}\looseness=-1

\section{Experiments}\label{sec:dep_results}

We conduct experiments on the English Penn Treebank \cite[PTB;][]{marcus-etal-1993-building}, the Chinese Penn Treebank \cite[CTB;][]{ctb}, and the Universal Dependencies 2.2 \cite[UD2.2;][]{ud2.2}.
For UD2.2, we adopt the pseudo-projective transformation \cite{nivre-nilsson-2005-pseudo} to convert non-projective trees into projective trees following previous work \cite{wang-tu-2020-second,yang-tu-2022-headed}.
We report dataset statistics in \cref{app:datasets} and hyperparameter settings in \cref{app:hyperpara}.\looseness=-1

\paragraph{Accuracy.} We train the hexatagger model based on XLNet \cite{xlnet} and report the results on PTB and CTB in \cref{tab:ptb_results}. Furthermore, we evaluate hexatagger in a set of 12 topologically diverse languages on UD corpus, where we use Multilingual BERT \cite{bert} as the underlying model (see \cref{tab:ud}).
In PTB, we observe that hexatagger achieves state-of-the-art results, compared to models with custom architectures and even in some cases with extra annotation. In CTB and UD, hexatagger follows the best performance closely. 

\input{tabs/efficiency.tex}

\paragraph{Efficiency.}
We compare the efficiency of hexatagger with biaffine modules,\footnote{By biaffine model we refer to a slim parameterization of a dependency parser that scores the existence of a dependency between $w_i$ and $w_j$ using a biaffine attention layer over the words' contextualized representations.} which are the backbone of many neural graph-based parsers \cite{kiperwasser-goldberg-2016-simple,dozat-biaffine,mrini-etal-2020-rethinking,yang-tu-2022-headed}. 
As depicted in \cref{tab:efficiency_comparison}, we observe that our hexatagger is an order of magnitude faster and consumes less memory.
Further analysis is included in \cref{app:analysis}.

\section{Conclusion}
In summary, hexatagging, our novel scheme, offers a parallelizable and efficiently decodable backbone for dependency parsing. 
Without relying on custom architecture for dependency parsing, the hexatagger achieves state-of-the-art accuracy on several datasets using no more than 
a pretrained language model and linear classifiers.

\section*{Limitations}

\paragraph{Non-projectivity.} The primary theoretical limitation of hexatagger is that it can only produce projective dependency trees. 
We would like to explore the possibility of extending hexatagger to non-projective parsing for future work. 

\paragraph{Interpretibility.} As a trade-off for efficiency, hexatagger does not model dependency arcs directly. Compared to graph-based models that explicitly score arc scores between pairs of words, it is more difficult to interpret the output of hexatagger.\looseness=-1

\section*{Ethics Statement}
We do not believe the work presented here further amplifies biases already present in the datasets.
Therefore, we foresee no ethical concerns in this work.\looseness=-1

\section*{Acknowledgments}
We would like to thank Tim Vieira for his invaluable feedback throughout the process of this paper. Afra Amini is supported by ETH AI Center doctoral fellowship.

\bibliography{anthology,custom}
\bibliographystyle{acl_natbib}

\clearpage
\appendix

\section{Algorithms}
\input{algs/dep2bht.tex}
\input{algs/bht2dep.tex}

\section{Related Work} \label{sec:related-works}
Traditionally, approaches to dependency parsing have been taxonomized into graph-based and transition-based parsers. 
The authors of this paper take the stance that this distinction is misleading because the difference lies not in the models themselves, but rather in whether exact or approximate inference algorithms are employed. 
For instance, \citet{kuhlmann-etal-2011-dynamic} gives exact algorithms for transition-based dependency parsers, which exposes the inability to formally distinguish graph-based and transition-based parsers.
Thus, we classify our related work into sections: exact and approximate decoding. Further, we review works on tagging-based parsing which is the most relevant line of work to this paper.
\paragraph{Exact Decoding.} 
Most exact algorithms for projective dependency parsing models apply a modified form of the CKY algorithm on nested dependency trees.
The best runtime among the commonly deployed algorithms  $\bigo{N^3}$ \citep{eisner-1996-acl}, but algorithms based on fast matrix multiplication exist and can achieve a lower runtime bound \cite{cohen-parsing}.
However, exact decoding of \emph{non-projective parsers} is intractable unless under independence assumptions, e.g., edge factored assumption \citep{mcdonald-satta-2007-complexity}.
Edge-factored parsers \citep{mcdonald-etal-2005-non, dozat-etal-2017-stanfords} construct graphs by scoring all possible arcs between each pair of words. They then use the maximum spanning tree (MST) finding algorithms for decoding to build the valid dependency trees with maximum score in $\bigo{N^2}$ \citep{zmigrod-etal-2020-please}.
The discussed algorithms are exact in inferring the dependency structure, however, they are neither fast nor parallelizable.

\paragraph{Approximate Decoding.} 
Despite not being exact, transition-based parsers offer faster and typically linear-time parsing algorithms \citep{kudo-matsumoto-2002-japanese, yamada-matsumoto-2003-statistical, nivre-2003-efficient}.
The dependency tree is inferred with a greedy search through transition system actions. Following this approach, actions are not predicted in parallel and the configuration of the transition system (stack and buffer) needs to be modeled with a neural network \citep{chen-manning-2014-fast}, which prevents using pretrained models out of the box.

\paragraph{Tagging-based parsing.} 
Inspired by \citeposs{bangalore-joshi-1999-supertagging} seminal work \emph{supertagging}, a recent line of work aims to utilize pretrained models and parse dependency trees by inferring tags for each word in the input sequence. 
\citet{li-etal-2018-seq2seq, kiperwasser-ballesteros-2018-scheduled} predict the relative position of the dependent with respect to its parent as the tag. They then use beam tree constraints \citep{lee-etal-2016-global} to infer valid dependency trees. \citet{strzyz-etal-2019-viable} provides a framework for analyzing similar tagging schemes. Although these works have demonstrated potential in this area, none achieved state-of-the-art results compared to custom architectures and algorithms developed for dependency parsing. Additionally, the output space, or size of the tag set, is unrestricted, which limits the efficiency of this approach.

\section{Analysis} \label{app:analysis}
\paragraph{\textsc{Left-first} vs. \textsc{Right-first}.} \label{app:left-right}
We examine the effect of the two orders of binarization of \cref{alg:dep2bht} in \cref{tab:left_right_first}. In our experiments, the choice of left-first or right-first order has little to no effect on parsing performance. 
\input{tabs/left_right_first.tex}

\section{Efficiency Evaluation} \label{app:efficiency}
For efficiency comparison, we use BERT-large as the base feature encoder for both \Hexataggertt{} and \texttt{Biaffine}.
We use the English PTB test set and truncate or pad the input sentences to the control length. 
The results are averaged over 3 random runs on the same server with one Nvidia A100-80GB GPU. The other experimental settings are kept the same (i.e., the version of PyTorch and Transformer, FP32 precision, batching).

\section{Datasets}\label{app:datasets}

\paragraph{Preprocessing.} Following previous work \cite{kiperwasser-goldberg-2016-simple,dozat-biaffine}, the dependency annotations are derived by the Stanford Dependency converter v3.3.0 \cite{de2008stanford} from the treebank annotations. 
Punctuation is omitted for evaluation. 
Gold part-of-speech tags are provided to the model both during training and evaluation following the code released by \citet{mrini-etal-2020-rethinking}.

Some other authors use system-predicted part-of-speech tags \cite{zhou-zhao-2019-head} or use mixed configurations. 
E.g., \citet{yang-tu-2022-headed} uses gold part-of-speech tags on CTB and UD, while not using any on PTB, \citet{dozat-biaffine} uses gold part-of-speech tags on CTB but system-predicted ones on PTB.
Our preliminary experiments show that removing the usage of part-of-speech information barely affects the UAS metric, and gives us a performance of 97.4 UAS and 95.8 LAS on PTB.

\paragraph{Splits.} All the datasets splits are consistent with previous work. For PTB, we follow the standard split of \citet{marcus-etal-1993-building}, resulting in 39,832 sentences for training, 1,700 for development, and 2,416 for testing. For CTB, we follow the split of \citet{zhang-clark-2008-tale}, resulting in 16,091 sentences for training, 803 for development, and 1,910 for testing. For UD2.2, we follow \citet{yang-tu-2022-headed} and use the standard splits of the following corpora for experiments:  BG-btb, CA-ancora, CS-pdt, DE-gsd, EN-ewt, ES-ancora, FR-gsd, IT-isdt, NL-alpino, NO-rrt, RO-rrt, RU-syntagrus.

\paragraph{Licenses.} The PTB and CTB datasets are licensed under LDC User Agreement. The UD2.2 dataset is licensed under the Universal Dependencies License Agreement.

\section{Hyperparameter Settings}\label{app:hyperpara}

We use the Python NLTK package to process the datasets, i.e., converting CoNLL-U formatted data to dependency trees, extracting dependency arcs from dependency trees for evaluation, implementing \cref{alg:dep2bht,alg:bht2dep}. 
For UD, we apply MaltParser v1.9.2\footnote{\url{http://www.maltparser.org/download.html}} to pseudo-projectivize the non-projective trees \cite{nivre-nilsson-2005-pseudo}.

We use xlnet-large-cased\footnote{\url{https://huggingface.co/xlnet-large-cased}} for English PTB, chinese-xlnet-mid\footnote{\url{https://huggingface.co/hfl/chinese-xlnet-mid}} for CTB, and bert-multilingual-cased\footnote{\url{https://huggingface.co/bert-base-multilingual-cased}} for UD.

The dimension of POS tag embedding is set to 256 for all experiments. On top of concatenated pretrained representations and POS embedding, we use a 3-layer BiLSTM with a hidden size of 768 for base-sized models (bert-multilingual-cased on UD) and 1024 for large-sized models (xlnet-large-cased on PTB and chinese-xlnet-mid on CTB).

Dropout layers with a rate of 0.33 are applied after the concatenated embedding layer, between LSTM layers, and before the MLP projection layer to hexatags. 

For training, we used AdamW with a learning rate of $2\mathrm{e}{-5}$ for pretrained LMs and $1\mathrm{e}{-4}$ for POS embedding, BiLSTM, and MLP. The gradient clipping threshold is set to $1.0$. The batch size is set to $32$.

\end{document}

%% file: figs/dep2tree.tex
\newcommand{\colsepdep}{0.4ex}
\newcommand{\arcangledep}{30}
\newcommand{\baselinedep}{{(-0.7ex)}}

\newcommand{\depfull}[0]{\hspace{-5pt}%
\begin{dependency}[baseline=\baselinedep,x=0.05\linewidth, y=0.05\linewidth,arc edge, arc angle=\arcangledep, text only label, label style={above}]
\begin{deptext}[column sep=\colsepdep]
A  \& B \& C \& D \\
\end{deptext}
\depedge{2}{1}{}
\depedge[]{2}{4}{}
\depedge{4}{3}{}
\end{dependency}\hspace{-6pt}
}

\newcommand{\depfullblue}[0]{\hspace{-5pt}%
\begin{dependency}[baseline=\baselinedep,x=0.05\linewidth, y=0.05\linewidth, arc edge,arc angle=\arcangledep, text only label, label style={above}]
\begin{deptext}[column sep=\colsepdep]
A  \& B \& C \& D \\
\end{deptext}
\depedge{2}{1}{}
\depedge[blue]{2}{4}{}
\depedge{4}{3}{}
\end{dependency}\hspace{-6pt}
}

\newcommand{\depdobluer}[2]{\hspace{-5pt}%
\begin{dependency}[baseline=\baselinedep,x=0.05\linewidth, y=0.05\linewidth,arc edge, arc angle=\arcangledep, text only label, label style={above}]
\begin{deptext}[column sep=\colsepdep]
{\color{blue} #1}  \& {\color{blue} #2} \\
\end{deptext}
\depedge[blue]{1}{2}{}
\end{dependency}\hspace{-6pt}
}

\newcommand{\depdor}[2]{\hspace{-5pt}%
\begin{dependency}[baseline=\baselinedep,x=0.05\linewidth, y=0.05\linewidth,arc edge, arc angle=\arcangledep, text only label, label style={above}]
\begin{deptext}[column sep=\colsepdep]
#1  \& #2 \\
\end{deptext}
\depedge{1}{2}{}
\end{dependency}\hspace{-6pt}
}

\newcommand{\depdobluel}[2]{\hspace{-5pt}%
\begin{dependency}[baseline=\baselinedep,x=0.05\linewidth, y=0.05\linewidth,arc edge, arc angle=\arcangledep, text only label, label style={above}]
\begin{deptext}[column sep=\colsepdep]
{\color{blue} #1}  \& {\color{blue} #2} \\
\end{deptext}
\depedge[blue]{2}{1}{}
\end{dependency}\hspace{-6pt}
}

\newcommand{\depdol}[2]{\hspace{-5pt}%
\begin{dependency}[baseline=\baselinedep,x=0.05\linewidth, y=0.05\linewidth,arc edge, arc angle=\arcangledep, text only label, label style={above}]
\begin{deptext}[column sep=\colsepdep]
#1  \& #2 \\
\end{deptext}
\depedge{2}{1}{}
\end{dependency}\hspace{-6pt}
}

\newcommand{\depb}[0]{%
\small
\begin{tikzpicture}[x=0.05\linewidth, y=0.05\linewidth][%
\draw[] (2.5, 4) node[inner sep=1pt, blue] (N1) {B};

\draw[] (1, 2.5) node[inner sep=1pt] (NA) {A};

\draw[] (5.5, 2.5) node[inner sep=1pt] (NB) {D};

\draw[] (4, 1) node[inner sep=1pt] (NC) {C};

\draw[thick, ->] (N1) -- (NA);
\draw[thick, ->] (N1) -- (NB);
\draw[thick, ->] (NB) -- (NC);
\end{tikzpicture}}

\newcommand{\depa}[0]{%
\small
\begin{tikzpicture}[x=0.05\linewidth, y=0.05\linewidth][%
\draw[] (2.5, 4) node[inner sep=1pt] (N1) {B};

\draw[] (1, 2.5) node[inner sep=1pt, blue] (NA) {A};

\draw[] (5.5, 2.5) node[inner sep=1pt] (NB) {D};

\draw[] (4, 1) node[inner sep=1pt] (NC) {C};

\draw[thick, ->] (N1) -- (NA);
\draw[thick, ->] (N1) -- (NB);
\draw[thick, ->] (NB) -- (NC);
\end{tikzpicture}}

\newcommand{\depab}[0]{%
\small
\begin{tikzpicture}[x=0.05\linewidth, y=0.05\linewidth][%
\draw[] (2.5, 4) node[inner sep=1pt, blue] (N1) {B};

\draw[] (1, 2.5) node[inner sep=1pt, blue] (NA) {A};

\draw[] (5.5, 2.5) node[inner sep=1pt] (NB) {D};

\draw[] (4, 1) node[inner sep=1pt] (NC) {C};

\draw[thick, blue, ->] (N1) -- (NA);
\draw[thick, ->] (N1) -- (NB);
\draw[thick, ->] (NB) -- (NC);
\end{tikzpicture}}

\newcommand{\depd}[0]{%
\small
\begin{tikzpicture}[x=0.05\linewidth, y=0.05\linewidth][%
\draw[] (2.5, 4) node[inner sep=1pt] (N1) {B};

\draw[] (1, 2.5) node[inner sep=1pt] (NA) {A};

\draw[] (5.5, 2.5) node[inner sep=1pt, blue] (NB) {D};

\draw[] (4, 1) node[inner sep=1pt] (NC) {C};

\draw[thick, ->] (N1) -- (NA);
\draw[thick, ->] (N1) -- (NB);
\draw[thick, ->] (NB) -- (NC);
\end{tikzpicture}}

\newcommand{\depc}[0]{%
\small
\begin{tikzpicture}[x=0.05\linewidth, y=0.05\linewidth][%
\draw[] (2.5, 4) node[inner sep=1pt] (N1) {B};

\draw[] (1, 2.5) node[inner sep=1pt] (NA) {A};

\draw[] (5.5, 2.5) node[inner sep=1pt] (NB) {D};

\draw[] (4, 1) node[inner sep=1pt, blue] (NC) {C};

\draw[thick, ->] (N1) -- (NA);
\draw[thick, ->] (N1) -- (NB);
\draw[thick, ->] (NB) -- (NC);
\end{tikzpicture}}

\newcommand{\depdc}[0]{%
\small
\begin{tikzpicture}[x=0.05\linewidth, y=0.05\linewidth][%
\draw[] (2.5, 4) node[inner sep=1pt] (N1) {B};

\draw[] (1, 2.5) node[inner sep=1pt] (NA) {A};

\draw[] (5.5, 2.5) node[inner sep=1pt, blue] (NB) {D};

\draw[] (4, 1) node[inner sep=1pt, blue] (NC) {C};

\draw[thick, ->] (N1) -- (NA);
\draw[thick, ->] (N1) -- (NB);
\draw[thick, blue, ->] (NB) -- (NC);
\end{tikzpicture}}

\newcommand{\depbd}[0]{%
\small
\begin{tikzpicture}[x=0.05\linewidth, y=0.05\linewidth][%
\draw[] (2.5, 4) node[inner sep=1pt, blue] (N1) {B};

\draw[] (1, 2.5) node[inner sep=1pt] (NA) {A};

\draw[] (5.5, 2.5) node[inner sep=1pt, blue] (NB) {D};

\draw[] (4, 1) node[inner sep=1pt] (NC) {C};

\draw[thick, ->] (N1) -- (NA);
\draw[thick, blue, ->] (N1) -- (NB);
\draw[thick, ->] (NB) -- (NC);
\end{tikzpicture}}

\newcommand{\baselinebht}{-0.3ex}

\newcommand{\rgilas}[2]{%
\small
\begin{tikzpicture}[baseline=\baselinebht, x=0.05\linewidth, y=0.05\linewidth][%
\draw[] (1.5, 2.) node[shape=circle,draw,thick,blue,inner sep=1pt] (N1) {R};

\draw[] (0.5, 0.3) node[shape=rectangle,draw,thick, inner sep=3pt] (NA) {#1};

\draw[] (2.7, 0.3) node[shape=rectangle,draw,thick, inner sep=3pt] (NB) {#2};

\draw[thick, -] (N1) -- (NA);
\draw[thick, -] (N1) -- (NB);
\end{tikzpicture}}

\newcommand{\dogilas}[0]{%
\small
\begin{tikzpicture}[baseline=\baselinebht, x=0.05\linewidth, y=0.05\linewidth][%
\draw[] (1.5, 1.5) node[shape=circle,draw,thick,inner sep=1pt] (N1) {R};
\draw[] (2.7, 0.3) node[shape=rectangle,draw,thick,inner sep=2pt] (N2) {B};

\draw[] (3.6, 2.7) node[shape=circle,draw,blue,thick,inner sep=2pt] (N4) {L};

\draw[] (0.5, 0.3) node[shape=rectangle,draw,thick,inner sep=2pt] (NA) {A};
\draw[] (5.7, 1.5) node[shape=circle, draw,thick,inner sep=1pt] (NB) {R};
\draw[] (4.8, 0.3) node[shape=rectangle,draw,thick,inner sep=2pt] (NC) {C};
\draw[] (6.6, 0.3) node[shape=rectangle,draw,thick,inner sep=2pt] (Nn) {D};

\draw[thick] (NA) -- (N1);
\draw[thick] (N1) -- (N4);
\draw[thick] (N2) -- (N1);
\draw[thick] (NC) -- (NB);
\draw[thick] (NB) -- (N4);
\draw[thick] (NB) -- (Nn);
\end{tikzpicture}}



\newcommand{\newstacka}[1]{%
\small
\begin{tikzpicture}[x=0.05\linewidth, y=0.05\linewidth][%
\draw[] (1.5, 1.5) node[shape=circle,thick,inner sep=0pt,blue] (N1) {\tt \$};

\draw[] (0.3, 0.5) node[shape=rectangle,blue,draw,thick,inner sep=2pt] (NA) {#1};

\draw[thick,blue] (NA) -- (N1);
\end{tikzpicture}}

\newcommand{\newdepa}[0]{%
\begin{dependency}[baseline=-1ex,x=0.05\linewidth, y=0.05\linewidth,arc edge, arc angle=\arcangledep, text only label, label style={above}]
\begin{deptext}[column sep=\colsepdep]
A  \\
\end{deptext}
\end{dependency}}


\newcommand{\newstackaa}[0]{%
\small
\begin{tikzpicture}[x=0.05\linewidth, y=0.05\linewidth][%
\draw[] (1.5, 1.5) node[shape=circle,blue,draw,thick,inner sep=1pt] (N1) {R};
\draw[] (2.7, 0.3) node[blue,inner sep=1pt] (N2) {$\varnothing$};

\draw[] (2.5, 2.5) node[blue,inner sep=0pt] (N4) {\tt \$};

\draw[] (0.5, 0.3) node[shape=rectangle,draw,thick,inner sep=2pt] (NA) {A};

\draw[thick] (NA) -- (N1);
\draw[thick,blue] (N1) -- (N4);
\draw[thick,blue] (N2) -- (N1);
\end{tikzpicture}}

\newcommand{\newdepaa}[0]{%
\begin{dependency}[baseline=-1ex,x=0.05\linewidth, y=0.05\linewidth,arc edge, arc angle=\arcangledep, text only label, label style={above}, edge horizontal padding=2pt]
\begin{deptext}[column sep=\colsepdep]
A  \\
\end{deptext}
\deproot[<-, edge height=4ex,blue]{1}{}
\end{dependency}}


\newcommand{\newstackaab}[0]{%
\small
\begin{tikzpicture}[x=0.05\linewidth, y=0.05\linewidth][%
\draw[] (1.5, 1.5) node[shape=circle,draw,thick,inner sep=1pt] (N1) {R};
\draw[] (2.7, 0.3) node[shape=rectangle,blue,draw,thick,inner sep=2pt] (N2) {B};

\draw[] (2.5, 2.5) node[inner sep=0pt] (N4) {\tt \$};

\draw[] (0.5, 0.3) node[shape=rectangle,draw,thick,inner sep=2pt] (NA) {A};

\draw[thick] (NA) -- (N1);
\draw[thick,blue] (N1) -- (N4);
\draw[thick,blue] (N2) -- (N1);
\end{tikzpicture}}

\newcommand{\newdepaab}[0]{%
\begin{dependency}[baseline=-1ex,x=0.05\linewidth, y=0.05\linewidth,arc edge, arc angle=\arcangledep, text only label, label style={above}, edge horizontal padding=2pt]
\begin{deptext}[column sep=\colsepdep]
{\color{blue} A}  \& {\color{blue} B} \\
\end{deptext}
\depedge[blue]{1}{2}{}
\end{dependency}}


\newcommand{\newstackaabb}[0]{%
\small
\begin{tikzpicture}[x=0.05\linewidth, y=0.05\linewidth][%
\draw[] (1.5, 1.5) node[shape=circle,draw,thick,inner sep=1pt] (N1) {R};
\draw[] (2.7, 0.3) node[shape=rectangle,draw,thick,inner sep=2pt] (N2) {B};

\draw[] (2.7, 2.7) node[shape=circle,blue,draw,thick,inner sep=2pt] (N4) {L};
\draw[] (3.9, 3.9) node[blue,inner sep=0pt] (N5) {\tt  \$};
\draw[] (3.9, 1.5) node[blue,inner sep=0pt] (N6) {$\varnothing$};

\draw[] (0.5, 0.3) node[shape=rectangle,draw,thick,inner sep=2pt] (NA) {A};

\draw[thick] (NA) -- (N1);
\draw[thick] (N1) -- (N4);
\draw[thick] (N2) -- (N1);
\draw[thick,blue] (N4) -- (N5);
\draw[thick,blue] (N6) -- (N4);
\end{tikzpicture}}

\newcommand{\newdepaabb}[0]{%
\begin{dependency}[baseline=-1ex,x=0.05\linewidth, y=0.05\linewidth,arc edge, arc angle=\arcangledep, text only label, label style={above}, edge horizontal padding=2pt]
\begin{deptext}[column sep=\colsepdep]
A  \&  B  \\
\end{deptext}
\depedge{1}{2}{}
\end{dependency}}

\newcommand{\newstackaabbb}[0]{%
\small
\begin{tikzpicture}[x=0.05\linewidth, y=0.05\linewidth][%
\draw[] (1.5, 1.5) node[shape=circle,draw,thick,inner sep=1pt] (N1) {R};
\draw[] (2.7, 0.3) node[shape=rectangle,draw,thick,inner sep=2pt] (N2) {B};

\draw[] (2.7, 2.7) node[shape=circle,draw,thick,inner sep=2pt] (N4) {L};
\draw[] (3.9, 3.9) node[inner sep=0pt] (N5) {\tt  \$};
\draw[] (3.9, 1.5) node[inner sep=0pt] (N6) {$\varnothing$};

\draw[] (0.5, 0.3) node[shape=rectangle,draw,thick,inner sep=2pt] (NA) {A};

\draw[thick] (NA) -- (N1);
\draw[thick] (N1) -- (N4);
\draw[thick] (N2) -- (N1);
\draw[thick] (N4) -- (N5);
\draw[thick] (N6) -- (N4);
\end{tikzpicture}}


\newcommand{\newstackaabbc}[0]{%
\small
\begin{tikzpicture}[x=0.05\linewidth, y=0.05\linewidth][%
\draw[] (1.5, 1.5) node[shape=circle,draw,thick,inner sep=1pt] (N1) {R};
\draw[] (2.7, 0.3) node[shape=rectangle,draw,thick,inner sep=2pt] (N2) {B};

\draw[] (3.6, 2.7) node[shape=circle,draw,thick,inner sep=2pt] (N4) {L};
\draw[] (4.9, 3.9) node[inner sep=0pt] (N5) {\tt  \$};

\draw[] (0.5, 0.3) node[shape=rectangle,draw,thick,inner sep=2pt] (NA) {A};
\draw[] (5.7, 1.5) node[shape=circle,blue, draw,thick,inner sep=1pt] (NB) {R};
\draw[] (4.8, 0.3) node[shape=rectangle,blue,draw,thick,inner sep=2pt] (NC) {C};
\draw[] (6.6, 0.3) node[blue, inner sep=0pt] (Nn) {$\varnothing$};

\draw[thick] (NA) -- (N1);
\draw[thick] (N1) -- (N4);
\draw[thick] (N2) -- (N1);
\draw[thick] (N4) -- (N5);
\draw[thick,blue] (NC) -- (NB);
\draw[thick, blue] (NB) -- (N4);
\draw[thick, blue] (NB) -- (Nn);
\end{tikzpicture}}

\newcommand{\newdepaabbc}[0]{%
\begin{dependency}[baseline=-1ex,x=0.05\linewidth, y=0.05\linewidth,arc edge, arc angle=\arcangledep, text only label, label style={above}, edge horizontal padding=2pt]
\begin{deptext}[column sep=\colsepdep]
A  \& B \& {\color{blue} C} \\
\end{deptext}
\depedge{1}{2}{}
\depedge[blue]{3}{1}{}
\end{dependency}}


\newcommand{\dashrule}[1][black]{%
  \color{#1}\rule[\dimexpr.5ex-.2pt]{4pt}{.4pt}\xleaders\hbox{\rule{4pt}{0pt}\rule[\dimexpr.5ex-.2pt]{4pt}{.4pt}}\hfill\kern0pt%
}
\newcommand{\mydashrule}[0]{%
\\[-2\jot]
\multicolumn{4}{@{}c@{}}{\makebox[\linewidth]{\dashrule[black!60]}} \\[-\jot]}

\newcommand{\thickdashrule}[1][black]{%
  \color{#1}\rule[\dimexpr.5ex-.2pt]{4pt}{1.2pt}\xleaders\hbox{\rule{4pt}{0pt}\rule[\dimexpr.5ex-.2pt]{4pt}{1.2pt}}\hfill\kern0pt%
}
\newcommand{\mythickdashrule}[0]{%
\\[-2\jot]
\multicolumn{4}{@{}c@{}}{\makebox[\linewidth]{\thickdashrule[black!60]}} \\[-\jot]}

\newcommand{\newstackaabbcd}[0]{%
\small
\begin{tikzpicture}[x=0.05\linewidth, y=0.05\linewidth][%
\draw[] (1.5, 1.5) node[shape=circle,draw,thick,inner sep=1pt] (N1) {R};
\draw[] (2.7, 0.3) node[shape=rectangle,draw,thick,inner sep=2pt] (N2) {B};

\draw[] (3.6, 2.7) node[shape=circle,draw,thick,inner sep=2pt] (N4) {L};
\draw[] (4.9, 3.9) node[inner sep=0pt] (N5) {\tt  \$};

\draw[] (0.5, 0.3) node[shape=rectangle,draw,thick,inner sep=2pt] (NA) {A};
\draw[] (5.7, 1.5) node[shape=circle, draw,thick,inner sep=1pt] (NB) {R};
\draw[] (4.8, 0.3) node[shape=rectangle,draw,thick,inner sep=2pt] (NC) {C};
\draw[] (6.6, 0.3) node[shape=rectangle,blue,draw,thick,inner sep=2pt] (Nn) {D};

\draw[thick] (NA) -- (N1);
\draw[thick] (N1) -- (N4);
\draw[thick] (N2) -- (N1);
\draw[thick] (N4) -- (N5);
\draw[thick] (NC) -- (NB);
\draw[thick] (NB) -- (N4);
\draw[thick, blue] (NB) -- (Nn);
\end{tikzpicture}}

\newcommand{\bhta}[0]{%
\small
\begin{tikzpicture}[x=0.05\linewidth, y=0.05\linewidth][%
\draw[] (1.5, 1.5) node[shape=circle,draw,thick,inner sep=1pt] (N1) {R};
\draw[] (2.7, 0.3) node[shape=rectangle,draw,thick,inner sep=2pt] (N2) {B};

\draw[] (3.6, 2.7) node[shape=circle,draw,thick,inner sep=2pt] (N4) {L};
\draw[] (4.9, 3.9) node[inner sep=0pt] (N5) {\tt  \$};

\draw[] (0.5, 0.3) node[shape=rectangle,draw,thick,blue, inner sep=2pt] (NA) {A};
\draw[] (5.7, 1.5) node[shape=circle, draw,thick,inner sep=1pt] (NB) {R};
\draw[] (4.8, 0.3) node[shape=rectangle,draw,thick,inner sep=2pt] (NC) {C};
\draw[] (6.6, 0.3) node[shape=rectangle,draw,thick,inner sep=2pt] (Nn) {D};

\draw[thick] (NA) -- (N1);
\draw[thick] (N1) -- (N4);
\draw[thick] (N2) -- (N1);
\draw[thick] (N4) -- (N5);
\draw[thick] (NC) -- (NB);
\draw[thick] (NB) -- (N4);
\draw[thick] (NB) -- (Nn);
\end{tikzpicture}}

\newcommand{\bhtb}[0]{%
\small
\begin{tikzpicture}[x=0.05\linewidth, y=0.05\linewidth][%
\draw[] (1.5, 1.5) node[shape=circle,draw,thick,inner sep=1pt] (N1) {R};
\draw[] (2.7, 0.3) node[shape=rectangle,draw,blue, thick,inner sep=2pt] (N2) {B};

\draw[] (3.6, 2.7) node[shape=circle,draw,thick,inner sep=2pt] (N4) {L};
\draw[] (4.9, 3.9) node[inner sep=0pt] (N5) {\tt  \$};

\draw[] (0.5, 0.3) node[shape=rectangle,draw,thick,blue,inner sep=2pt] (NA) {A};
\draw[] (5.7, 1.5) node[shape=circle, draw,thick,inner sep=1pt] (NB) {R};
\draw[] (4.8, 0.3) node[shape=rectangle,draw,thick,inner sep=2pt] (NC) {C};
\draw[] (6.6, 0.3) node[shape=rectangle,draw,thick,inner sep=2pt] (Nn) {D};

\draw[thick] (NA) -- (N1);
\draw[thick] (N1) -- (N4);
\draw[thick] (N2) -- (N1);
\draw[thick] (N4) -- (N5);
\draw[thick] (NC) -- (NB);
\draw[thick] (NB) -- (N4);
\draw[thick] (NB) -- (Nn);
\end{tikzpicture}}

\newcommand{\bhtR}[0]{%
\small
\begin{tikzpicture}[x=0.05\linewidth, y=0.05\linewidth][%
\draw[] (1.5, 1.5) node[shape=circle,draw,thick,blue, inner sep=1pt] (N1) {R};
\draw[] (2.7, 0.3) node[shape=rectangle,draw,blue, thick,inner sep=2pt] (N2) {B};

\draw[] (3.6, 2.7) node[shape=circle,draw,thick,inner sep=2pt] (N4) {L};
\draw[] (4.9, 3.9) node[inner sep=0pt] (N5) {\tt  \$};

\draw[] (0.5, 0.3) node[shape=rectangle,draw,thick,blue, inner sep=2pt] (NA) {A};
\draw[] (5.7, 1.5) node[shape=circle, draw,thick,inner sep=1pt] (NB) {R};
\draw[] (4.8, 0.3) node[shape=rectangle,draw,thick,inner sep=2pt] (NC) {C};
\draw[] (6.6, 0.3) node[shape=rectangle,draw,thick,inner sep=2pt] (Nn) {D};

\draw[thick] (NA) -- (N1);
\draw[thick] (N1) -- (N4);
\draw[thick] (N2) -- (N1);
\draw[thick] (N4) -- (N5);
\draw[thick] (NC) -- (NB);
\draw[thick] (NB) -- (N4);
\draw[thick] (NB) -- (Nn);
\end{tikzpicture}}

\newcommand{\bhtc}[0]{%
\small
\begin{tikzpicture}[x=0.05\linewidth, y=0.05\linewidth][%
\draw[] (1.5, 1.5) node[shape=circle,draw,thick,inner sep=1pt] (N1) {R};
\draw[] (2.7, 0.3) node[shape=rectangle,draw,thick,inner sep=2pt] (N2) {B};

\draw[] (3.6, 2.7) node[shape=circle,draw,thick,inner sep=2pt] (N4) {L};
\draw[] (4.9, 3.9) node[inner sep=0pt] (N5) {\tt  \$};

\draw[] (0.5, 0.3) node[shape=rectangle,draw,thick,inner sep=2pt] (NA) {A};
\draw[] (5.7, 1.5) node[shape=circle, draw,thick,inner sep=1pt] (NB) {R};
\draw[] (4.8, 0.3) node[shape=rectangle,draw,thick,blue, inner sep=2pt] (NC) {C};
\draw[] (6.6, 0.3) node[shape=rectangle,draw,thick,inner sep=2pt] (Nn) {D};

\draw[thick] (NA) -- (N1);
\draw[thick] (N1) -- (N4);
\draw[thick] (N2) -- (N1);
\draw[thick] (N4) -- (N5);
\draw[thick] (NC) -- (NB);
\draw[thick] (NB) -- (N4);
\draw[thick] (NB) -- (Nn);
\end{tikzpicture}}

\newcommand{\bhtd}[0]{%
\small
\begin{tikzpicture}[x=0.05\linewidth, y=0.05\linewidth][%
\draw[] (1.5, 1.5) node[shape=circle,draw,thick,inner sep=1pt] (N1) {R};
\draw[] (2.7, 0.3) node[shape=rectangle,draw,thick,inner sep=2pt] (N2) {B};

\draw[] (3.6, 2.7) node[shape=circle,draw,thick,inner sep=2pt] (N4) {L};
\draw[] (4.9, 3.9) node[inner sep=0pt] (N5) {\tt  \$};

\draw[] (0.5, 0.3) node[shape=rectangle,draw,thick,inner sep=2pt] (NA) {A};
\draw[] (5.7, 1.5) node[shape=circle, draw,thick,inner sep=1pt] (NB) {R};
\draw[] (4.8, 0.3) node[shape=rectangle,draw,thick,blue, inner sep=2pt] (NC) {C};
\draw[] (6.6, 0.3) node[shape=rectangle,draw,thick,blue, inner sep=2pt] (Nn) {D};

\draw[thick] (NA) -- (N1);
\draw[thick] (N1) -- (N4);
\draw[thick] (N2) -- (N1);
\draw[thick] (N4) -- (N5);
\draw[thick] (NC) -- (NB);
\draw[thick] (NB) -- (N4);
\draw[thick] (NB) -- (Nn);
\end{tikzpicture}}

\newcommand{\bhtRR}[0]{%
\small
\begin{tikzpicture}[x=0.05\linewidth, y=0.05\linewidth][%
\draw[] (1.5, 1.5) node[shape=circle,draw,thick,inner sep=1pt] (N1) {R};
\draw[] (2.7, 0.3) node[shape=rectangle,draw,thick,inner sep=2pt] (N2) {B};

\draw[] (3.6, 2.7) node[shape=circle,draw,thick,inner sep=2pt] (N4) {L};
\draw[] (4.9, 3.9) node[inner sep=0pt] (N5) {\tt  \$};

\draw[] (0.5, 0.3) node[shape=rectangle,draw,thick,inner sep=2pt] (NA) {A};
\draw[] (5.7, 1.5) node[shape=circle, draw,thick,blue, inner sep=1pt] (NB) {R};
\draw[] (4.8, 0.3) node[shape=rectangle,draw,thick,blue, inner sep=2pt] (NC) {C};
\draw[] (6.6, 0.3) node[shape=rectangle,draw,thick,blue, inner sep=2pt] (Nn) {D};

\draw[thick] (NA) -- (N1);
\draw[thick] (N1) -- (N4);
\draw[thick] (N2) -- (N1);
\draw[thick] (N4) -- (N5);
\draw[thick] (NC) -- (NB);
\draw[thick] (NB) -- (N4);
\draw[thick] (NB) -- (Nn);
\end{tikzpicture}}

\newcommand{\bhtL}[0]{%
\small
\begin{tikzpicture}[x=0.05\linewidth, y=0.05\linewidth][%
\draw[] (1.5, 1.5) node[shape=circle,draw,thick,inner sep=1pt] (N1) {R};
\draw[] (2.7, 0.3) node[shape=rectangle,draw,thick,inner sep=2pt] (N2) {B};

\draw[] (3.6, 2.7) node[shape=circle,draw,thick,blue, inner sep=2pt] (N4) {L};
\draw[] (4.9, 3.9) node[inner sep=0pt] (N5) {\tt  \$};

\draw[] (0.5, 0.3) node[shape=rectangle,draw,thick,inner sep=2pt] (NA) {A};
\draw[] (5.7, 1.5) node[shape=circle, draw,thick,inner sep=1pt] (NB) {R};
\draw[] (4.8, 0.3) node[shape=rectangle,draw,thick, inner sep=2pt] (NC) {C};
\draw[] (6.6, 0.3) node[shape=rectangle,draw,thick,inner sep=2pt] (Nn) {D};

\draw[thick] (NA) -- (N1);
\draw[thick] (N1) -- (N4);
\draw[thick] (N2) -- (N1);
\draw[thick] (N4) -- (N5);
\draw[thick] (NC) -- (NB);
\draw[thick] (NB) -- (N4);
\draw[thick] (NB) -- (Nn);
\end{tikzpicture}}

%% file: tabs/ud.tex
\begin{table*}\small
 \setlength{\abovecaptionskip}{-4pt}
 \setlength{\belowcaptionskip}{-4pt}
	\centering
	{ \setlength{\tabcolsep}{.8em}
		\makebox[\linewidth]{\resizebox{\linewidth}{!}{%
    \begin{tabular}{lccccccccccccc}
        \toprule
            & 	bg   & 	 ca   &  cs   &  de   &  en  &  es  &  fr   &  it  & nl  & no  & ro  & ru   &  Avg. \\ \toprule
        \citet{zhang-etal-2020-efficient}   &  90.77   &  91.29   &  91.54   &  80.46   &  87.32   &  90.86   &  87.96   &  91.91   &  88.62   &  91.02   &  86.90   &  93.33   &  89.33 \\
        \citet{wang-tu-2020-second}   &  90.53   &  92.83   &  92.12   &  81.73   &  89.72   &  92.07   &  88.53   &  92.78   &  90.19   &  91.88   &  85.88   &  92.67   &  90.07   \\ 	\midrule
    \multicolumn{13}{c}{+$\text{BERT}_{\text{multilingual}}$}  \\  \midrule 
        \citet{wang-tu-2020-second}   &  {91.30}   &  93.60   &  92.09   &  82.00   &  90.75   &  92.62   &  89.32   &  93.66   &  91.21   &  91.74   &  86.40   &  92.61    &  90.61 \\ 
        \citet{dozat-biaffine}   &  90.30   &  \textbf{94.49}   &  {92.65}   &  \textbf{85.98}   &  91.13   &  93.78   &  \textbf{91.77}   & 94.72   & 91.04   & 94.21   & 87.24   &  \textbf{94.53}   & 91.82 \\ 
         \citet{yang-tu-2022-headed}   &  91.10   &  94.46   & 92.57   &  85.87   &  \textbf{91.32}   &  \textbf{93.84}   &  91.69   &  \textbf{94.78}   &  {91.65}   &  {94.28}   &  {87.48}   & 94.45   &  \textbf{91.96} \\  \midrule
         \Hexataggertt   &  \textbf{92.87}   &  93.79   &  \textbf{92.82}   &  85.18   &  90.85   &  93.17   &  91.50   &  94.72   &  \textbf{91.89}   &  {93.95}   &  \textbf{87.54}   &  94.03   &  {91.86}
         \\ \bottomrule	\\		
\end{tabular}}}
 }
	\caption{LAS scores on the test sets of 12 languages in UD 2.2. Hexatagger achieves competitive performance in all languages and is state-of-the-art in 4 languages.}
	\label{tab:ud} 
	\vskip -.12in
\end{table*}

%% file: tabs/ptb.tex
\begin{table}
\centering
 \setlength{\belowcaptionskip}{-10pt}
\resizebox{\columnwidth}{!}{%
\begin{tabular}{@{}cccccc@{}}\toprule
  & \multicolumn{2}{c}{PTB} & \multicolumn{2}{c}{CTB} & \\
\cmidrule(lr){2-3} \cmidrule(lr){4-5}
Model  & UAS & LAS & UAS & LAS\\ \midrule
\citet{zhou-zhao-2019-head}$^*$         & 97.0  & 95.4  & 91.2  & 89.2 \\
\citet{mrini-etal-2020-rethinking}$^*$  & 97.4  & {96.3}& 94.6  & 89.3\\ \midrule
\citet{chen-manning-2014-fast}          & 91.8  & 89.6  & 83.9  & 82.4 \\
\citet{dozat-biaffine}                  & 95.7  & 94.1  & 89.3  & 88.2 \\
\citet{yang-tu-2022-headed}$^{\#}$         & 97.4  & 95.8  & \textbf{93.5}  & \textbf{92.5}    \\  \midrule
\Hexataggertt                       & \textbf{97.4}  & \textbf{96.4}  & 93.2  &  91.9    \\ \bottomrule
\end{tabular}}
\caption{Results on PTB and CTB. $^{*}$ indicates usage of extra constituency annotation. $^{\#}$ is our re-implementation using the same pretrained encoder with hexatagger.}
\label{tab:ptb_results}
\vskip -.12in
\end{table}

%% file: tabs/efficiency.tex
\begin{table}
\centering
 \setlength{\belowcaptionskip}{-10pt}
\resizebox{\columnwidth}{!}{%
\begin{tabular}{@{}ccccc@{}}\toprule
 & \multicolumn{2}{c}{Speed (sent/s) $\uparrow$} & \multicolumn{2}{c}{Memory (GB) $\downarrow$} \\
\cmidrule(lr){2-3}
\cmidrule(lr){4-5}
Sent length & \Hexataggertt & \Biaffinett & \Hexataggertt & \Biaffinett \\ \midrule
32 & 2916 & 493 & 2.9 & 4.5 \\
64 & 3011 & 328 & 3.0 &  10.1 \\
128 & 2649 & 202 & 3.7 &  30.6 \\
256 & 3270 & 98 & 4.5 &  56.2$^\star$ \\ \midrule
 {overall} & \textbf{3176} & 338 & \textbf{3.0} &  10.6 \\ \bottomrule
\end{tabular}}
\caption{Comparison of parsing speed and memory consumption on PTB test set. Results are averaged over 3 random runs on the same server with one Nvidia A100-80GB GPU using BERT-large as encoder. We use a batch size of 128 sentences, except for $^\star$ that uses 64, which otherwise results in an out-of-memory error.}
\label{tab:efficiency_comparison}
\end{table}

%% file: algs/dep2bht.tex
\begin{algorithm}[!h]
\footnotesize
\caption{Create a BHT from a dependency tree.} 
\begin{algorithmic}[1]
    \Procedure{\textsc{dep2tree}}{$\texttt{head}$}
    \State stack.push(\texttt{head}) 
    \For{{\tt dep} in \textsc{LeftDep}({\tt head})}
    \State \textsc{dep2tree}({\tt dep})
    \State {\tt left} $\gets$ stack.pop()
    \State {\tt right} $\gets$ stack.pop()
    \State stack.push(\textsc{MakeNode}(\circledr{}, {\tt left}, {\tt right}))
    \EndFor
    \For{{\tt dep} in \textsc{RightDep}({\tt head})}
    \State \textsc{dep2tree}({\tt dep})
    \State {\tt left} $\gets$ stack.pop()
    \State {\tt right} $\gets$ stack.pop()
    \State stack.push(\textsc{MakeNode}(\circledl{}, {\tt left}, {\tt right}))
    \EndFor
    \EndProcedure
\end{algorithmic}\label{alg:dep2bht}
\end{algorithm}

%% file: algs/bht2dep.tex
\begin{algorithm}[!h]
\footnotesize
\caption{Create a dependency tree from a BHT.}
\begin{algorithmic}[1]
    \Procedure{\textsc{tree2dep}}{$\texttt{node}$}
    \If{{\tt node} is leaf}
    \State \Return {\tt node}
    \EndIf
    \State {\tt left} $\gets$ \textsc{tree2dep}({\tt node}.left)
    \State {\tt right} $\gets$ \textsc{tree2dep}({\tt node}.right)
    \If{{\tt node} is \circledl{}}
    \vspace{-2ex}
    \State \Return \textsc{MakeArc}(\depdor{{\tt left}}{{\tt right}})
    \EndIf
    \vspace{-2ex}
    \State \Return \textsc{MakeArc}(\depdol{{\tt left}}{{\tt right}})
    \EndProcedure
\end{algorithmic}\label{alg:bht2dep}
\end{algorithm}

%% file: tabs/left_right_first.tex
\begin{table}[ht]
\centering
\resizebox{0.85\columnwidth}{!}{%
\begin{tabular}{@{}cccccc@{}}\toprule
  & \multicolumn{2}{c}{PTB} & \multicolumn{2}{c}{CTB} & \\
\cmidrule(lr){2-3} \cmidrule(lr){4-5}
Model  & UAS & LAS & UAS & LAS\\ \midrule
Right-first & 97.2  & 96.3  & \textbf{93.2}  &  91.9    \\ 
Left-first & \textbf{97.4}   &  \textbf{96.4} & 93.1  &  91.9    \\ \bottomrule
\end{tabular}}
\caption{Comparison of left-first and right-first binarization.}
\label{tab:left_right_first}
\end{table}